# Video-Based Facial Expression Recognition Using Local Directional Binary Pattern

*Sahar Hooshmand, Ali Jamali Avilaq, Amir Hossein Rezaie*
*Electrical Engineering Dept., AmirKabir Univarsity of Technology*
*Tehran, Iran*

*Abstract*—Automatic facial expression analysis is a challenging issue and influenced so many areas such as human computer interaction. Due to the uncertainties of the light intensity and light direction, the face gray shades are uneven and the expression recognition rate under simple Local Binary Pattern is not ideal and promising. In this paper we propose two state-of-the-art descriptors for person-independent facial expression recognition. First the face regions of the whole images in a video sequence are modeled with Volume Local Directional Binary pattern (VLDBP), which is an extended version of the LDBP operator, incorporating movement and appearance together. To make the survey computationally simple and easy to expand, only the co-occurrences of the Local Directional Binary Pattern on three orthogonal planes (LDBP-TOP) are debated. After extracting the feature vectors the K-Nearest Neighbor classifier was used to recognize the expressions. The proposed methods are applied to the videos of the Extended Cohn-Kanade database (CK+) and the experimental outcomes demonstrate that the offered techniques achieve more accuracy in comparison with the classic and traditional algorithms.

*Keywords- Chi-Square Statistics; Facial Expression Recognition; Feature Vector; K-Nearest Neighbor Classifier; Local Directional Binary Pattern*

## I. Introduction

Facial expression recognition is one of the most controversial topics in computer vision, since it brings in a new dimension of human-computer interaction. Although, much growth has been achieved [1,2,3], but due to the facial expressions' variability, a high accuracy recognition isn't easily attained.

There are two prevalent procedures to extract facial features: geometric feature-based methods and appearance-based methods [4]. Geometric features offer the shape and locations of facial components, which are extracted to form a feature vector that indicates the face geometry. Appearance-based methods utilize some image filters such as Gabor wavelets, and they are involved with either the whole-face or specific face-regions to extract the appearance changes of the face.

In this work, we use the Extended Cohn-Kanade facial expression database (CK+) [5], which is consist of 593 sequences across 123 subjects which are FACS coded at the peak frame. All sequences are from the neutral face to the peak expression. The last frame represents one of the basic moods which are listed as below:
Anger, Contempt, Disgust, Fear, Happiness, Sadness and Surprise.

It should be noted that facial expression recognition has two important aspects: feature extraction and classifier designing. In this paper we experimentally provide a survey on Local Binary Pattern (LBP) features [1], [6,7,8] for facial expression recognition and we compare it with a newer method by the name of Local Directional Binary Pattern (LDBP) [9]. Then we discuss about two novel methods, VLDBP and LDBP-TOP which are the extended form of LDBP for video processing. Local Binary Pattern was first described in 1994 [10]. It has since been found to be a powerful feature for texture classification; and recently have been introduced to represent faces in facial images' analysis. The most important factors about this pattern are the computational simplicity, robustness against illumination changes and head rotations [1]. Much progress has been made in the last decade and Local Binary Patterns improved so much. We are going to have a brief review on these improvements and compare the result of recognition rates on different methods which are based on LBP and LDBP.

As the second vital aspect of facial expression recognition, we used a K-nearest neighbor classifier to determine the expression. K-nearest neighbor (KNN) is an instance-based classification method, firstly introduced by Cover and Hart [11]. The KNN classification algorithm tries to find the "K" nearest neighbors of the current sample and uses a majority vote to determine the class label of the current sample [12]. It should be noted that if insufficient features are used, even the best classifiers would face failure.

The remainder of this paper is structured as follows: We present a brief review of related work in the next section. The Extended Cohn-Kanade database is introduced in Section III. Facial expression analysis by Single image processing is discussed in Section IV. The main contribution of this paper is appeared in section V, VI and VII. The first two parts discuss about facial expression recognition using VLDBP and LDBP-TOP features and Section VII explains the classification method using K-nearest neighbor classifier. Also the priority of these novel procedures over some other LBP-based methods is debated. Finally, Section VIII concludes the paper.

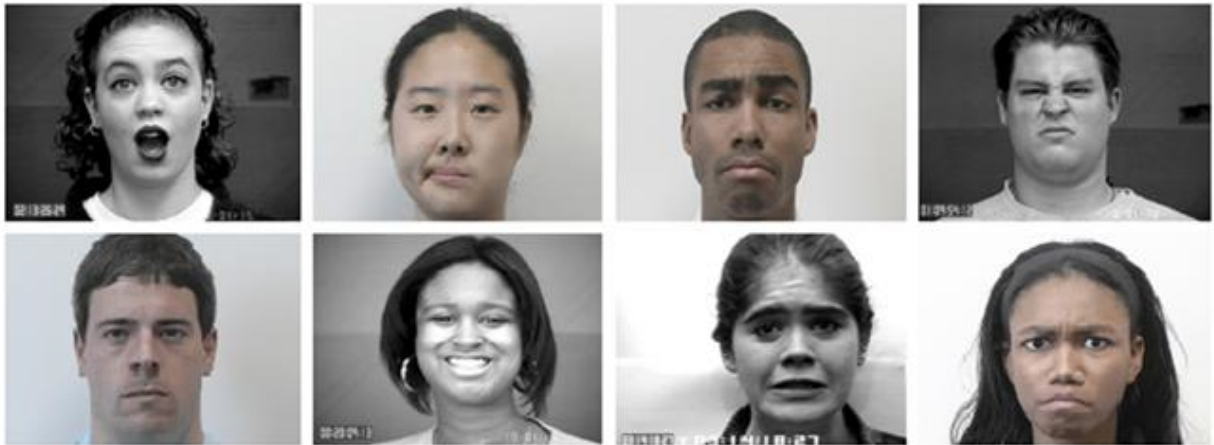

Figure 1. The sample facial expression images from the Extended Cohn–Kanade database.

## II. RELATED WORK

There has been considerable effort toward solving the problem of Facial expression recognition (FER). In human to human interaction, it has been discovered that verbal hints provide 7% of the meaning of the message; vocal cues, 38%; and facial expressions, 55%. Thus facial expression provides more information about the interaction than the spoken words [13]. Due to the various applications, FER makes many researchers be interested in this field. Much improvement has been made through the last decade and here we briefly review some previous work in order to put our work in context. In some existing work the appearance changes of faces are modeled. General spatial analysis including Linear Discriminate Analysis (LDA) [14], and Gabor wavelet analysis [15] have been applied to privileged face regions to extract the facial appearance changes. Among these methods, Gabor wavelet analysis is widely used in facial expression recognition due to its superior performance [16] but it's actually memory and time consuming.

In some other work [2], Facial geometry analysis has been widely applied on facial components to extract their shapes and locations. In image sequences, the facial motions can be determined by stacking up the geometrical displacement of facial feature points between the current frame and the primary frame.

Another method to recognize facial expression is using action units. Facial Action Coding System (FACS) is a system to taxonomies human facial movements by their appearance on the face. Due to subjectivity and time consumption issues, FACS has been established as a computed automated system that detects faces in videos, extracts the geometrical features of the faces, and then produces temporal profiles of each facial movement. Recently, Valstar and Pantic [17] presented a fully automatic AU detection system that can automatically concentrate on facial points in the very first frame and identify AU temporal segments using a proper subset of most erudite spatiotemporal features selected by AdaBoost. However, the geometric feature-based method generally requires precise facial feature detection and a trustworthy tracking, which is difficult to embed in many situations.

Recently Local Binary Patterns have been introduced as effective appearance features for facial image analysis. In this work we are going to compare our novel methods by different LBP-based methods because LBP has raised great methods for its simple calculation process and stronger anti-interference ability in comparison with the methods mentioned in this part.

Several methods have been used to classify facial expressions such as Support Vector Machine (SVM) [18] and Bayesian Network (BN) [19]. There have been several intentions to track and recognize facial expressions based on optical flow analysis. Tian, Kanade and Cohn presented a Neural Network to recognize facial action units in image sequences [20].

Hidden Markov Models (HMMs) have been generally used to model the temporal behaviors of facial expressions from image sequences [21]. In 2010, Schmidt, Schels and Scheweker proposed an HMM classifier for facial expression recognition in image sequences [22]. But HMMs can't deal with dependencies in observation. Recently Le An, Kafai and Bhanu propose a Dynamic Bayesian Network (DBN) to merge the information from different cameras as well as the temporal indication from frames in a video sequence [23]. The advantage of using DBN is that if features from one camera were not extracted due to image capture failure, this information can still be inferred by DBN and, therefore, recognition may not fail.

## III. EXTENDED COHN-KANADE DATABASE (CK+)

The Cohn-Kanade (CK) database was published for the idea of promoting research into automatically detecting person-independent facial expressions in 2000 [24]. Since then, the CK database has become one of the most commonly used datasets. But some limitations were obvious, such as lack of the Standard protocols for most of the databases of that time. To solve such concerns, Kanade et al. presented the Extended Cohn-Kanade (CK+) database [5].

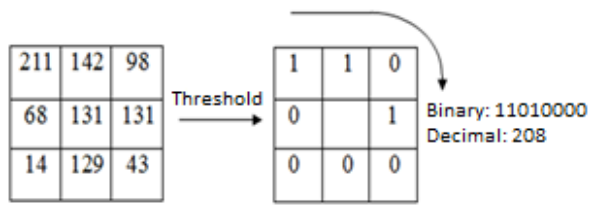

Figure 2. The basic LBP operator

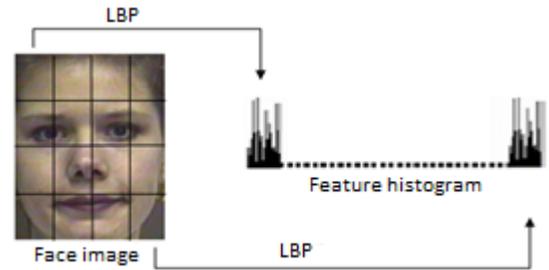

Figure 3. A face images is divided into small regions and a histogram is computed for each area. Then all the histograms are concatenated.

Participants were 18 to 50 years of age, 69% female, 81%, Euro-American, 13% Afro-American, and 6% other groups. The number of sequences is increased by 22% and the number of subjects by 27%. The video files are 593 sequences across 123 subjects. All sequences are from the neutral face to the peak expression and they represent one of the basic moods which are Anger, Contempt, Disgust, Fear, Happy, Sadness and Surprise. The target expression for each sequence is fully FACS coded and emotion labels have been modified and accredited. In addition to this, non-posed sequences for various types of smiles and their metadata have been included. Examples of facial expressions in CK+ database are given in Fig. 1.

## IV. FACIAL EXPRESSION ANALYSIS BY SINGLE IMAGE PROCESSING

One of the most common and effective methods in pattern recognition is using a simple algorithm by the name of Local Binary Pattern. This operator tags the pixels of a gray-scale image by getting done over a 3×3 neighborhood of each pixel with a central value and bring up a result as a binary number (refer to Fig. 2).Then a 256-bin histogram of the LBP labels would be computed [1].The limitation of the basic LBP operator is its small 3×3 neighborhood (9 gray values) which can not capture premiere features with large scale instructions. So it's important to crop the images to select the face parts. The Operator LBP (P, R) produces $2^P$ different values as outputs, corresponding to the $2^P$ different binary patterns that can be figured out by the P pixels in the neighbor area. Using circular neighborhoods, allows us to have any radius and number of pixels in the neighborhood. Therefore, there are different kinds of extended LBP which encompass several values for R and P.

$$LBP(x_c, y_c) = \sum_{k=1}^{k=7} S(g_i - g_c) 2^i$$

$$S(x) = \begin{cases} 1 & x \geq 0 \\ 0 & x < 0 \end{cases} \quad (1)$$

Where $g_c$ denotes the intensity value of the center pixel $(x_c, y_c)$, $g_i$ (i = 0, 1,..., 7) are the gray values of the surrounding eight pixels. It should be mentioned that to consider the local information of face components, images will be divided into small regions (Fig. 3) and a histogram will be computed for each area. Then all the histograms will be concatenated [6]. The LBP histogram contains information about the distribution of the local micro-patterns, such as edges, spots and flat areas, over the whole image, so can be used to statistically describe image characteristics [6]. Face images can be seen as the combinations of micro-patterns which can be effectively demonstrated by the LBP histograms. Referring Fig. 4, you will notice the micro-patterns which are shown by a central pixel and eight neighboring points. An LBP histogram computed over the whole face portion of an image, encodes only the incidences of the micro-patterns without any hint about their locations. As a matter of fact, the simple LBP can not identify the rotation of the black and white circles in several micro-patterns. By rotating the neighboring points around the central pixel, the LBP code of the micro-pattern will remain the same. To solve this problem, we are going to use an algorithm by the name of Local Directional Binary Pattern (LDBP), which is introduced in [9] in 2014. This code has much better performance in comparison with the simple LBP. Another important subject to discuss is that, most of the facial expression recognition algorithms are based on the single image analysis. The operation of these methods outstandingly depends on illumination variations (gray-scale changes), head rotations or translation. Therefore, the video-based facial expression recognition has attracted much attention in recent years. This kind of recognition utilizes the information of all video frames and provides more robustness against the problems mentioned above. Generally, the video-based analysis, results a higher recognition rate. The difference between a single image and video frames is that, the video is extended in a spatiotemporal domain. So the motion and appearance are combined. Hence the main contribution of this paper is to develop the effective LDBP algorithm to utilize it in video processing. The next two parts are about the novel algorithms we applied on videos of CK+ Database.

## V. VOLUME LOCAL DIRECTIONAL BINARY PATTERN (VLDBP)

As you noticed in previous sections, we have to introduce an effective algorithm which can process the facial expressions by high accuracy rate. For this purpose, we extend the Local Directional Binary Pattern [9], and we reach to a novel method which has an acceptable recognition rate. A VLDBP histogram computed over the whole face image of a video frame, encodes the incidences of the micro-patterns with a hint about the rotation of neighboring points around the central pixel. VLDBP is a binary code which is assigned to each pixel of an input face image. The pattern is calculated by comparing the relative edge values of a pixel ($m_i$ (i = 0, 1 . . . 7)) in different directions. So, by considering the directions, the pattern can detect the rotations in micro-patterns. We calculate eight directional edge response value of a particular pixel using Kirsch templates which are introduced in Fig. 5 in eight different orientations ($m_0$~$m_7$). It's obvious that the matrix $m_2$ is obtained by 90 degree rotating of $m_0$. Figure 6 shows the complete computing procedure for VLDBP (1, 4, 1). As we don't know the motion direction, we also select the neighboring points in a circle and not only in a direct line.

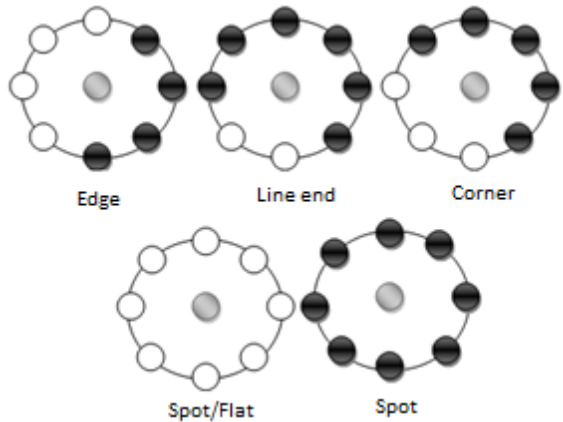

Figure 4. Examples of micro patterns (White circles represent ones & black circles represent zeroes)

We begin by making four non-overlapping patches on each sequence and sampling eight neighboring points for each pixel in all frames of the volume. Then, in element-by-element form, we multiply the sampling points and the Kirsch matrices. Then we sum them up to form the Kirsch values. It should be mentioned that, the central value is the median (or mean) of the other eight Kirsch values. In these matrices, we select the four neighboring points, and then every four points in the Kirsch values would be compared with the value of the central pixel of the middle frame to get a binary value. Finally, we produce the VLDBP code by multiplying the binary values with weights given to the corresponding pixel and we sum up the results. Finally we transform the binary number into decimal to mention the value of VLDBP (L, P, R). It should be mentioned that "P" is the number of local neighboring points in a circle by radius "R", around the central pixel in one frame. And actually the "L" is the time interval. In equation (2), "$K_i$" is the Kirsch value in previous, middle and posterior frames. These values are determined by colored cells. And "Kc" is the central value of the middle frame, which demonstrates the Kirsch values in figure 6.

$$VLDBP(L, P, R) = \sum_{i=0}^{3P+1} S(K_i - K_c)2^i$$

$$S(x) = \begin{cases} 1 & x \geq 0 \\ 0 & x < 0 \end{cases} \quad (2)$$

Let's imagine an X ×Y ×T video volume in which: ($x_c$ ∈{0,…., $X_c$-1}, $y_c$ ∈{0,…., $Y_c$-1}, $t_c$ ∈{0,…., $T_c$-1}). For calculating VLDBP (L, P, R) feature for this volume, the central part is only debated, because an adequately wide neighborhood can't be used on the borders in this 3D space. The VLDBP code is calculated for each pixel in every patch of the volume, and the distribution of the code is used as a feature vector, which is demonstrated by "D". The joint distribution of the gray levels is denoted as "v". Now we reach to a 256 bins histogram and it must be normalized to get a coherent explanation. VLDBP combines the movement and appearance to describe videos.

$$m_0 = \begin{bmatrix} -3 & -3 & 5 \\ -3 & 0 & 5 \\ -3 & -3 & 5 \end{bmatrix} \quad m_1 = \begin{bmatrix} -3 & 5 & 5 \\ -3 & 0 & 5 \\ -3 & -3 & -3 \end{bmatrix}$$

$$m_2 = \begin{bmatrix} 5 & 5 & 5 \\ -3 & 0 & -3 \\ -3 & -3 & -3 \end{bmatrix} = \text{rot90} (m_0) \quad m_3 = \begin{bmatrix} 5 & 5 & -3 \\ 5 & 0 & -3 \\ -3 & -3 & -3 \end{bmatrix} = \text{rot90} (m_1)$$

$$m_4 = \begin{bmatrix} 5 & -3 & -3 \\ 5 & 0 & -3 \\ 5 & -3 & -3 \end{bmatrix} = \text{rot90} (m_2) \quad m_5 = \begin{bmatrix} -3 & -3 & -3 \\ 5 & 0 & -3 \\ 5 & 5 & -3 \end{bmatrix} = \text{rot90} (m_3)$$

$$m_6 = \begin{bmatrix} -3 & -3 & -3 \\ -3 & 0 & -3 \\ 5 & 5 & 5 \end{bmatrix} = \text{rot90} (m_4) \quad m_7 = \begin{bmatrix} -3 & -3 & -3 \\ -3 & 0 & 5 \\ -3 & 5 & 5 \end{bmatrix} = \text{rot90} (m_5)$$

Figure 5. Kirsch operator template

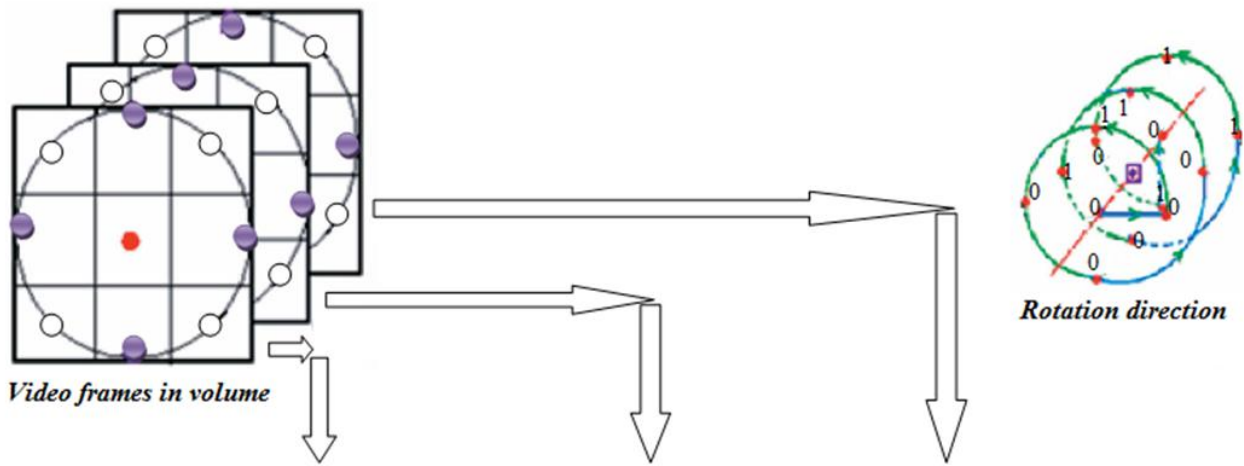

Figure 6. Procedure of VLDBP (1, 4, 1)

$$D = v(LDBP_{L,P,R}(x, y, t))$$
$$x \in \lceil R \rceil, X - 1 - \lceil R \rceil$$
$$y \in \lceil R \rceil, Y - 1 - \lceil R \rceil \quad (3)$$
$$t \in \lceil L \rceil, T - 1 - \lceil L \rceil$$

## VI. LOCAL DIRECTIONAL BINARY PATTERN FROM THREE ORTHOGONAL PLANES

In the proposed VLDBP, the parameter "P" specifies the number of features. A large "P" produces a long histogram. On the other hand a small "P" makes the feature vector shorter but also causes losing more information. As we mentioned before, the number of patterns for VLDBP is $2^{3P+2}$.

So when the number of neighboring points increases, the number of patterns will become very large, Due to this rapid growth, it is tough to extend VLDBP to have a large number of neighboring points and this restricts its applications. Likewise, the neighboring frames with a time interval less than "L" will be dropped. To solve these problems, we offer a simple method by concatenating LDBP on three orthogonal planes: XY, XT, and YT, discussing only the co-occurrences in these three planes (Shown in Figure 7(a)).

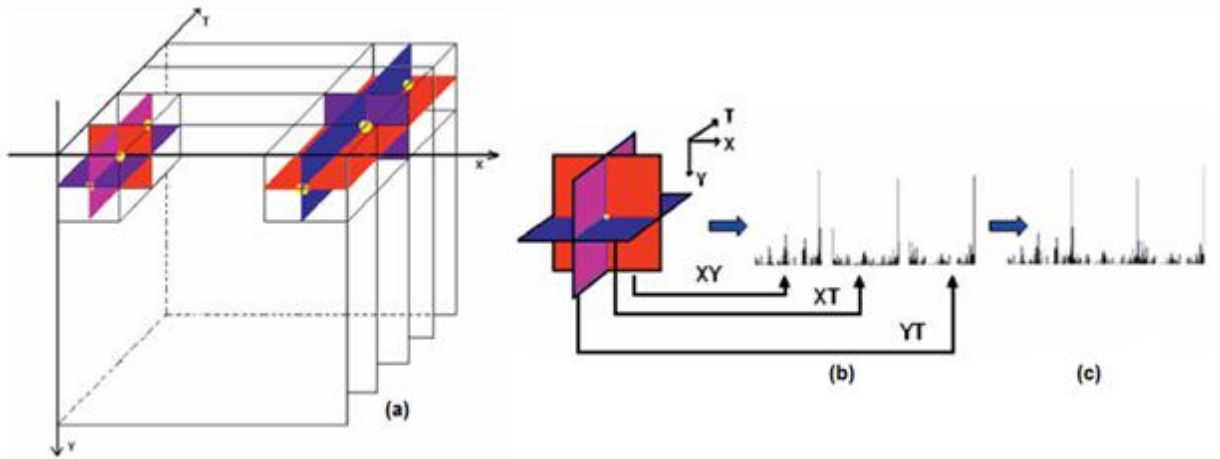

Figure 7. (a) Three orthogonal planes to extract neighboring points. (b) LDBP histogram from each plane. (c) Concatenated feature histogram

The XY plane includes the space information, whereas the XT and YT planes provide information about the space-time transitions. With this approach, the number of bins is only $3 \times 2^p$, much smaller than $2^{3P+2}$, as shown in figure 6, which provides a simpler propagation to many neighboring points and also decreases the complexity of computations. There are two principal differences between VLDBP and LDBP-TOP.

First, the VLDBP uses three parallel frames of which only the middle one contains the central pixel. On the other hand, the LDBP-TOP uses three orthogonal planes that cross in the central pixel. Second, VLDBP considers the co-occurrences of all neighboring points from three parallel frames, which causes to a long feature vector. LDBP-TOP collects the information from three separate planes and then concatenates them together (Shown in Fig. 9). So the feature vector would be much shorter when the number of neighboring points grows and the number of bins would be reasonable. Actually, the radii in axes X, Y, and T and the number of neighboring points in the XY, XT, and YT planes can be different, which can be marked as $R_X$, $R_Y$, and $R_T$.

But we consider the R=1 for all planes. Compared with VLDBP, not all the information of the video (image volume), but only the features from three planes are exerted. Figure 7(a) illustrates three orthogonal planes. Figure 7(b) shows the image histogram in the XY, XT and YT planes, The XY plane only contains the information about the space, whereas the XT plane, gave the perception of one row changing in time. On the other hand the YT plane explains the movement of one column in temporal space. The LDBP-TOP code is extracted from the XY, XT, and YT planes, which are defined as XY _ LDBP, XT _ LDBP, and YT _ LDBP for all pixels and then they would be concatenated into a single histogram, as you notice in figure7(c).

As you noticed in Fig. 8, to extract the LDBP-TOP feature, we begin with making an image volume. In this paper, we continue our survey by making four non-overlapping patches on each sequence. Then the LDBP features are extracted for each video on three orthogonal planes, which include the space information in XY plane and the spatiotemporal data on XT and YT planes. As you know, for extracting the LDBP feature [9], we begin by sampling eight neighboring points for each pixel in all frames of the volume and then, in element-by-element form, we multiply the sampling points and the Kirsch matrices. Then we sum them up to form the Kirsch values. The Kirsch values are going to be compared with the central value (median or mean of the eight Kirsch values) of the corresponding frame. We do this procedure in every three planes of the volume. Let's imagine an $X \times Y \times T$ video volume in which: (xc $\in \{0,...., Xc-1\}$, yc $\in \{0,...., Yc-1\}$, tc $\in \{0,...., Tc-1\}$). In calculating LDBP _ TOP $_{PXY ;PXT ;PYT ;RX;RY ;RT}$ feature for this volume, the central part is only debated, because an adequately wide neighborhood can't be used on the borders in this 3D space. The histogram of the volume can be explained as:

$$H_{i,j} = \sum_{x,y,t} I\{f_i(x,y,t) = i\}$$
$$i = 0,\ldots, n_j - 1 \; ; j = 0, 1, 2$$
(4)

In which $n_j$ is the number of different labels produced by the LDBP operator in the jth plane (j=0: XY, 1: XT, 2: YT), expresses the LDBP code of central pixel (x,y,t) in the jth

plane and $I\{A\} = \begin{cases} 1 & \text{if A is true} \\ 0 & \text{if A is false} \end{cases}$

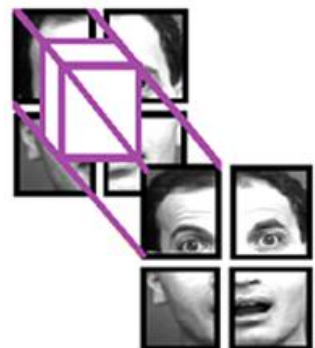

Figure 8. Facial expression representation- Making patches and extract features in each block volume.

TABLE I. COMPARISON BETWEEN SEVERAL LBP-BASED ALGORITHMS AND OUR NOVEL METHODS; USING THE DIFFERENT VALUES OF "K"

| methods | Recognition rate for different values of "K" | | | | | |
|---|---|---|---|---|---|---|
| | K=1 | K=3 | K=5 | K=10 | K=15 | K=18 |
| LDBP-TOP | 74.3 | 75.2 | 79.2 | 81.6 | 81.0 | 81.9 |
| VLDBP | 73.2 | 75.1 | 77.2 | 80.7 | 81.1 | 80.9 |
| CLBP-TOP | 72.2 | 74.4 | 76.8 | 80.0 | 80.2 | 80.3 |
| EVLBP | 70.9 | 71.3 | 77.2 | 79.2 | 79.8 | 78.2 |
| LBP-TOP | 68.7 | 68.9 | 71.4 | 74.0 | 72.6 | 71.0 |
| VLBP | 65.4 | 66.8 | 70.3 | 72.1 | 72.8 | 73.3 |

TABLE II. CONFUSION MATRIX OF 7-CLASS FACIAL EXPRESSION RECOGNITION; USING VLDBP FEATURES AND K-NEAREST NEIGHBOR CLASSIFIER (K=10)

| (%) | Surprise | Happiness | Sadness | Fear | Disgust | Contempt | Angry |
|---|---|---|---|---|---|---|---|
| Surprise | **85.1** | 0 | 0 | 14.9 | 0 | 0 | 0 |
| Happiness | 0 | **80.2** | 6.1 | 2.9 | 3.5 | 7.3 | 0 |
| Sadness | 6.8 | 2.7 | **72.9** | 4.1 | 0 | 0 | 13.5 |
| Fear | 1.2 | 23.2 | 2.8 | **65.1** | 4.2 | 3.5 | 0 |
| Disgust | 0 | 0 | 2.7 | 14.5 | **79.3** | 0 | 3.5 |
| Contempt | 0 | 19.3 | 0 | 3.8 | 1.4 | **70.7** | 2.8 |
| Angry | 0 | 0 | 14.3 | 0 | 6.2 | 4.5 | **75.5** |

TABLE III. CONFUSION MATRIX OF 7-CLASS FACIAL EXPRESSION RECOGNITION; USING LDBP-TOP FEATURES AND K-NEAREST NEIGHBOR CLASSIFIER (K=10)

| (%) | Surprise | Happiness | Sadness | Fear | Disgust | Contempt | Angry |
|---|---|---|---|---|---|---|---|
| Surprise | **86.9** | 0 | 0 | 13.1 | 0 | 0 | 0 |
| Happiness | 0 | **83.8** | 5.1 | 2.6 | 3.7 | 4.8 | 0 |
| Sadness | 7.1 | 2.4 | **74.6** | 3.8 | 0 | 0 | 12.1 |
| Fear | 0 | 21.5 | 2.9 | **68.0** | 4.8 | 2.8 | 0 |
| Disgust | 0 | 0 | 0 | 17.6 | **82.4** | 0 | 0 |
| Contempt | 0 | 16.3 | 0 | 6.5 | 0 | **73.1** | 4.1 |
| Angry | 0 | 0 | 13.2 | 0 | 4.6 | 3.6 | **78.6** |

The histograms must be normalized to get a coherent explanation:

$$N_{i,j} = \frac{H_{i,j}}{\sum_{k=0}^{n_j-1} H_{k,j}} \quad (5)$$

In this histogram, a description of video is effectively established based on LDBP from three orthogonal planes.

The features from the XY plane include information about the appearance, and, feature from the XT and YT planes that contain co-occurrence of movement in horizontal and vertical directions. These three histograms are concatenated to form a global explanation of a video volume with the spatiotemporal features.

### VII. VLDBP & LDBP-TOP DESCRIPTORS FOR FACIAL VIDEO ANALYSIS

In this section, we accomplish a person-independent facial expression recognition using LDBP features. First, the face portions of all images in a video volume were detected by Viola-jones algorithm and then they were cropped. Then we divided them into small regions from which VLDBP and LDBP-TOP histograms were extracted and concatenated into a single, spatiotemporally enhanced feature histogram. Some parameters can be optimized for better feature extraction. For example, one of them is the number of patches or divided regions. We selected the 256-bin VLDBP(L,P,R) = VLDBP(1,4,1) and LDBP _ TOP $_{PXY;PXT;PYT;RX;RY;RT}$ =LDBP-TOP$_{4;4;4;1;1;1}$ operators. After making a video volume, we divided the whole face images into 20×20 pixels regions, with four 10×10 patches, giving a good trade-off between recognition rate and feature vector length.
According to FACS codes folders of the Extended Cohn-Kanade dataset, we divided the whole database to seven different moods which are: Anger, Contempt, Disgust, Fear, Happy, Sadness and Surprise.

After extracting all feature vectors of the whole videos of database, a K-nearest neighbor classifier is used to match the input video with the closest video volumes and classifies the input video to in the related group based on the similarity with each mood. Following [25], we also selected the Chi-square statistic (χ2) as the dissimilarity criterion for histograms, where S and M are two VLDBP or LDBP-TOP histograms as mentioned in equation (6). The parameter "K" in K-nearest neighbor classification demonstrates the number of the videos which are the closest ones to the input video and are going to compare with it.

$$\chi^2(S,M) = \sum_i \frac{(S_i - M_i)^2}{S_i + M_i} \quad (6)$$

The best choice of "K" depends upon the data; generally, larger values of "K" reduce the effect of noise on the classification, but make boundaries between classes less distinct. The special case where the class is predicted to be the class of the closest training sample (K=1), is called the nearest neighbor algorithm. This classification achieved an excellent performance in this 7-class task for both VLDBP and LDBP-TOP algorithms.

We compared the results with that reported in [1], [7], [8], where the authors used VLBP, LBP-TOP, CLBP and EVLBP. As we can not directly compare these several methods, we selected the same values for common parameters and the same preprocessing steps for all methods. All together, the comparison in Table I, demonstrates that our novel methods using VLDBP and LDBP-TOP features provide the better performances confusion matrix of this 7-class recognition is demonstrated in Table II and Table III.

while using K=10. We can observe that some moods can be recognized with a high accuracy, but on the other hand, some of them are easily confused while recognizing. The pair sadness and anger is difficult to recognize even for a human beings. The distinction between happiness and contempt failed. Because these expressions have a similar mouth motion.

### VIII. CONCLISION

A newfound approach to facial expression recognition was presented. A Volume LDBP method was developed to merge the motion and appearance together. A simpler LDBP-TOP operator based on concatenated LDBP histograms calculated from three orthogonal planes was also offered, making it easier to extract co-occurrence features from a larger number of neighboring points.

Experiments on Extended Cohn-Kanade database with a comparison to the other methods' results prove that our method is efficient for facial expression recognition. By using the K-nearest neighbor Classifier and selecting the K=10, we reach to the rates of 80.7% using VLDBP and 81.6% using LDBP-TOP. We used these video-base methods because not only this kind of recognition is computationally simple but also ,utilizes the information of all video frames and provides more robustness against the problems such as illumination variations (gray-scale changes), head rotations or translation. Generally, in comparison with the single image processing, the video-based analysis results a higher recognition rate and guarantees a promising outcome for real-world applications. The results gained from VLDBP and LDBP-TOP are better than those established in earlier studies. Furthermore; no gray-scale normalization is needed due to applying our descriptors to the face images.

### ACKNOWLEDGMENT

The authors would like to thank Professors Jeffery F. Cohn and Takeo Kanade for use of the Extended Cohn–Kanade facial expression database. And we appreciate the helpful comments and suggestions of the anonymous reviewers.

### REFERENCES

[1] G. Zhao, M. Pietikäinen, "Dynamic Texture Recognition Using Local Binary Patterns with an Application to Facial Expressions," IEEE Transactions on Pattern Analysis and Machine Intelligence, vol. 29, June 2007, pp. 915-928, doi:10.1109/TPAMI.2007.1110.

[2] M. Pantic, I. Patras, "Dynamics of Facial Actions and Their Temporal Segments from Face Profile Image Sequences," Systems, Man, and Cybernetics, Part B: Cybernetics, IEEE Transactions, vol. 36, Apr. 2006, pp. 433-449, doi:10.1109/TSMCB.2005.859075.

[3] X. Lu, A. Jain, "Deformation Modeling for Robust 3D Face Matching," Pattern Analysis and Machine Intelligence, IEEE Transactions, vol. 30, Aug. 2008, pp. 1346-1357, doi:10.1109/TPAMI.2007.70784.

[4] Y. Tian, T. Kanade, J. Cohn, "Handbook of Face Recognition", Springer, 2005, pp. 247-277.

[5] P. Lucey, J. F. Cohn, T. Kanade, J. Saragih, Z. Ambadar, "The Extended Cohn-Kanade Dataset (CK+), a Complete Dataset for

Action Unit and Emotion-specified Expression," Computer Vision and Pattern Recognition Workshops (CVPRW), 2010 IEEE Computer Society Conference, June 2010, pp. 94-101, doi:10.1109/CVPRW.2010.5543262.

[6] C. Shan, Sh. Gong, P. W. Mcowan, "Facial Expression Recognition Based on Local Binary Patterns: A Comprehensive Study," Image and Vision Computing on Elsevier, vol. 27, May 2009, pp. 803-816, doi:10.1016/j.imavis.2008.08.005.

[7] Zh. Guo, L. Zhang, D. Zhang, "A Complete Modeling of Local Binary Pattern Operator for Texture Classification," Image Processing, IEEE Transactions, vol. 19, Mar. 2010, pp. 1657-1663, doi:10.1109/TIP.2010.2044957.

[8] A. Hadid, M. Pietikäinen, "Learning Personal Specific Facial Dynamics for Face Recognition from Videos," Springer Berlin Heidelberg, Analysis and Modeling of Faces and Gestures, Lecture Notes in Computer Science, vol. 4778, 2007, pp. 1-15, doi:10.1007/978-3-540-75690-3_1.

[9] Y. Wang, G. HE, "Expression Recognition Algorithm Based on Local Directional Binary Pattern," Green Computing and Communications (GreenCom), 2013 IEEE and Internet of Things (iThings/CPSCom), IEEE International Conference on and IEEE Cyber, Physical and Social Computing, Aug. 2013, pp. 1458-1462, doi:10.1109/GreenCom-iThings-CPSCom.2013.257.

[10] D. C. He, L. Wang, "Texture Unit, Texture Spectrum and Texture Analysis," Geoscience and Remote Sensing, IEEE Transactions on, vol. 28, July 1990, pp. 509-512, doi:10.1109/TGRS.1990.572934.

[11] T. Cover, P. Hart, "Nearest neighbor pattern classification," Information Theory, IEEE Transactions on, vol. 13, Jan. 1967, pp. 21-27, doi:10.1109/TIT.1967.1053964.

[12] S. Zhang, X. Zhao, B. lei, "Facial Expression Recognition Using Sparse Representation," WSEAS Transactions on Systems, vol. 11, 2012, pp. 440-452.

[13] S. Moore, R. Bowden, "Local Binary Pattern for Multi-view Facial Expression Recognition," Computer Vision and Image Understanding on Elsevier, vol. 115, Apr. 2011, pp. 541-558, doi:10.1016/j.cviu.2010.12.001.

[14] P. N. Belhumeur, J. P. Hespanha, D. J Kriegman, "Eigen faces vs. fisher faces: Recognition Using Class Specific Linear Projection," Pattern Analysis and Machine Intelligence, IEEE Transactions on, vol. 19, July 1997, pp. 711-720, doi:10.1109/34.598228.

[15] M. J. Lyons, J. Budynek, S. Akamatsu, "Automatic Classification of Single Facial Images," Pattern Analysis and Machine Intelligence, IEEE Transactions on, vol: 21, Dec. 1999, pp. 1357-1362, doi:10.1109/34.817413.

[16] M. S. Bartlett, G. Littlewort, M. Frank, C. Lainscsek, I. Fasel, J. Movellan, "Recognition Facial Expression: Machine Learning and Application to Spontaneous Behavior," Computer Vision and Pattern Recognition, 2005. CVPR 2005. IEEE Computer Society Conference on, vol. 2, June 2005, pp.568-573, doi:10.1109/CVPR.2005.297.

[17] M. Valstar, M. Pantic, "Fully Automatic Facial Action Unit Detection and Temporal Analysis," IEEE Computer Vision and Pattern Recognition Workshop, CVPRW '06. Conference on, June 2006, pp. 149, doi:10.1109/CVPRW.2006.85.

[18] Y. Lai, Y. Zhang, "Using SVM to Design Facial Expression Recognition for Shape and Texture Features," Machine Learning and Cybernetics (ICMLC), 2010 International Conference on, vol. 5, July 2010, pp. 2697-2704, doi:10.1109/ICMLC.2010.5580938.

[19] C. Shan, Sh. Gong, P.W. Mcowan, "Dynamic Facial Expression Recognition Using a Bayesian Temporal Manifold Model," British Machine Vision Conference, 2006.

[20] Y. Tian, T. Kanade, J. Cohn, "Recognizing Action Units for Facial Expression Analysis," Pattern Analysis and Machine Intelligence, IEEE Transactions on, vol. 23, Feb. 2001, pp. 97-115, doi:10.1109/34.908962.

[21] M. Yeasin, B. Bullot, R. Sharma, "Facial Expression to Level of Interests: A Spatiotemporal Approach," Computer Vision and Pattern Recognition, 2004. CVPR 2004. Proceedings of the 2004 IEEE Computer Society Conference on, vol. 2, June 2004, pp. II-922 - II-927, doi:10.1109/CVPR.2004.1315264.

[22] M. Schemidt, M. Schels, F. Scheweker, "A Hidden Markov Model Based Approach for Facial Expression Recognition in Image Sequences," Artificial Neural Networks in Pattern Recognition Lecture notes in Computer Science, vol. 5998, 2010, pp. 149-160, doi:10.1007/978-3-642-12159-3_14.

[23] L. An, M. Kafai, B. Bhanu, "Dynamic Bayesian Network for Unconstrained Face Recognition in Surveillance Camera Networks," Emerging and Selected Topics in Circuits and Systems, IEEE Journal on, vol. 3, Apr. 2013, pp. 155-164, doi:10.1109/JETCAS.2013.2256752.

[24] T. Kanade, J. F. Cohn, Y. Tian, "Comprehensive Database for Facial Expression Analysis," Automatic Face and Gesture Recognition, 2000. Proceedings. Fourth IEEE International Conference on, Mar. 2000, pp. 46-53, doi:10.1109/AFGR.2000.840611.

[25] T. Ahonen, A. Hadid, M. Pietikäinen, "Face Recognition With Local Binary Patterns," Computer Vision - ECCV 2004 Lecture Notes in Computer Science, vol. 3021, May 2004, pp. 469-481, doi:10.1007/978-3-540-24670-1_36.